\documentclass{article}

\PassOptionsToPackage{numbers, square}{natbib}
\usepackage[final]{neurips_2022}

\usepackage[utf8]{inputenc} 
\usepackage[T1]{fontenc}    
\usepackage{hyperref}       
\usepackage{url}            
\usepackage{booktabs}       
\usepackage{amsfonts}       
\usepackage{nicefrac}       
\usepackage{microtype}      
\usepackage{xcolor}         
\usepackage{times}
\usepackage{latexsym}
\usepackage{amsmath}
\usepackage{multirow}

\title{Using Selective Masking as a Bridge between Pre-training and Fine-tuning}

\author{%
  Tanish Lad\\
  IIIT Hyderabad\\
  \texttt{tanish.lad@research.iiit.ac.in} \\
  \And
  Himanshu Maheshwari \\
  Adobe Research\thanks{The author was at IIIT Hyderabad during the work.} \\
  \texttt{him.maheshwari1999@gmail.com} \\
  \And
  Shreyas Kottukkal \\
  IIIT Hyderabad \\
  \texttt{shreyas.shankar@students.iiit.ac.in} \\
  \And
  Radhika Mamidi \\
  IIIT Hyderabad \\
  \texttt{radhika.mamidi@iiit.ac.in} \\
}

\begin{document}

\maketitle

\begin{abstract}
Pre-training a language model and then fine-tuning it for downstream tasks has demonstrated state-of-the-art results for various NLP tasks. Pre-training is usually independent of the downstream task, and previous works have shown that this pre-training alone might not be sufficient to capture the task-specific nuances. We propose a way to tailor a pre-trained BERT model for the downstream task via task-specific masking before the standard supervised fine-tuning. For this, a word list is first collected specific to the task. For example, if the task is sentiment classification, we collect a small sample of words representing both positive and negative sentiments. Next, a word's importance for the task, called the word's task score, is measured using the word list. Each word is then assigned a probability of masking based on its task score. We experiment with different masking functions that assign the probability of masking based on the word's task score. The BERT model is further trained on MLM objective, where masking is done using the above strategy. Following this standard supervised fine-tuning is done for different downstream tasks. Results on these tasks show that the selective masking strategy outperforms random masking, indicating its effectiveness.
\end{abstract}

\section{Introduction}

Pre-trained language models have been instrumental in spurring many advancements across various NLP tasks \citep{yang2020xlnet,beltagy2019scibert,liu2019roberta}. Most of these works follow the paradigm of unsupervised pre-training on large general-domain corpora followed by fine-tuning for downstream tasks, following the success of BERT {\citep{devlin-etal-2019-bert}}.
While the pre-training and fine-tuning paradigm has proven to be highly successful, \citet{gururangan-etal-2020-dont} has shown that even large LMs struggle to capture the complexity of a single textual domain. Thus, they suggest domain-specific pre-training on task-relevant corpora for better downstream performance in that domain. Other works \citep{beltagy2019scibert,chalkidis-etal-2020-legal,huang-etal-2020-clinical,chakraborty-etal-2020-biomedbert,covidtwitterbert}  show improvements by pre-training BERT like models on huge in-domain corpora relevant to the downstream tasks.

BERT uses the Cloze task \cite{cloze} for their masked LM (MLM) pre-training procedure wherein 15\% of input tokens are masked from each sequence at random. \citet{clark2020electra} and \citet{gu2020train} offer alternatives to this random masking strategy. \citet{clark2020electra} argues that learning a random 15\% of the input sequence is inefficient and suggests a ``replaced token detection" pre-training task instead. 
\citet{gu2020train} proposes a selective-masking strategy, arguing that there are task-specific tokens that are more important to mask than other tokens. They added a task-guided selective masking pre-training between general pre-training and fine-tuning to learn the domain-specific and task-specific language patterns. This strategy makes use of a small in-domain dataset for selective masking.

In this work, we propose a novel way to tailor the BERT model for the downstream task via task-specific masking before the standard supervised fine-tuning. We take a pre-trained BERT model and tailor it for downstream tasks using task-specific selective masking on a small chunk of BookCorpus (\citet{bookcorpus}). This tailored model is then fine-tuned for downstream tasks. Our method includes a novel approach to find tokens important for the downstream task using a list of seed words relevant to the downstream task. The word list and word embeddings are used to compute a `task' score for each word, which is used to calculate the masking probability of the word. We experiment with different masking functions and show considerable value in such selective masking. Unlike \citet{clark2020electra}, we use selective masking to train our model. Compared to \citet{gu2020train}, we use only word-lists instead of in-domain data for selective masking. Since BERT also uses BookCorpus, we do not use any new corpus for training. Collecting a word list is an easier alternative when in-domain datasets are scarce.
We also extend this approach to a wider variety of downstream tasks than the classification tasks explored by \citet{gu2020train}.

Experimental results on different downstream tasks such as \textbf{sentiment analysis, hate speech classification, formality detection and NER on informal text} show that this methodology performs better than random masking. Thus our methodology is both effective and generalizable.

\section{Task Specific Masking}
In this section, we first describe the method to calculate the task score of the word. Following that, we describe different masking functions that use this task score to assign a probability to each word. The word is then masked with this probability.

\subsection{Calculating Task Specific Score of a Word}
We leverage the classification framework in \citet{niu-etal-2017-study}, who calculated the formality score of a word for machine translation. We begin with a set of seed words indicative of different classes in downstream tasks. For example, in sentiment analysis, we choose negative-sentiment and positive-sentiment words as two sets of word-lists. Here we assume that the classification is binary (although we experiment with multi-class sentiment classification too). For another example, let us consider a non-classification task like NER on the domain of informal texts. Here our first list represents in-domain vocabulary and contains words usually found in informal texts. Our second list contains words not found in that domain, and thus here, we use words found in the formal texts. 


Following \citet{niu-etal-2017-study}, a Support Vector Machine (SVM) model is trained by assigning scores of $0$ and $10$ for the two sets of words, and a separating hyperplane is learned between Word2Vec \cite{word2vec} vector space representations of the two classes of seed words. \citet{niu-etal-2017-study} reported the best performance using an SVM model with Word2Vec representations. We did some experiments with GloVe embeddings \cite{pennington-etal-2014-glove} and SVM as well but word2vec gave better results. 
Once the SVM model is trained, euclidean distance to this hyperplane is used to measure the given word's task-specific score.  


\subsection{Masking Functions}
As described earlier, our approach uses a masking function to compute the probability of masking a given word based on its task score (obtained from the SVM model described earlier). If the downstream task is classification, then we mask both the extremes.  For example, if the task is sentiment classification, the language model is required to understand both extremes, viz. positive sentiments and negative sentiments. So we assign a high probability to both extremes. If the task requires the model to learn only one extreme, say NER on the informal dataset, then in that case, we assign a high probability to only one extreme (i.e., the one representing informality). 
In the following functions, $s$ is the euclidean distance of the word from the SVM hyperplane. $\alpha$, $\beta$, and $\gamma$ are constants that were adjusted such that approximately 15\% of the tokens are masked.

\noindent\textbf{Masking Function 1: Step Function}
\[ 
            f(s) =
            \begin{cases} 
              1 & s\leq \alpha\;  or\;   s\geq 10 - \alpha \\
              0 & otherwise 
           \end{cases}
        \]

\noindent\textbf{Masking Function 2: Linear Function}         
       \[ f(s) = \frac{ max \,(\, s \,,\, 10-s \,) - \alpha }{\beta}\]
    
\noindent\textbf{Masking Function 3: Concave Up (exponential)}
        \[ 
            f(s) = \frac{e^{\alpha.max(s, 10 - s)} - e^\beta}{\gamma}
        \]
        
   The step function only masks words at the extreme and completely discards other words. However, \citet{inbook} has argued that the appearance of an opinion word in a sentence does not necessarily mean that the sentence expresses a positive or negative opinion. Similarly, \citet{pavlick-tetreault-2016-empirical} has argued that formality is not only expressed by the use of formal/informal words but also "neutral" words, punctuations, capitalization, paraphrasing, etc. So in the remaining masking functions, we assign a non-zero probability to non-extreme words.
    The linear function assigns each word a probability linearly proportional to its importance. The exponential function glorifies the values which are towards the extreme and exponentially decreases the values as we move towards the center. It assigns each word a probability exponentially proportional to its importance.

\section{Experiments}
We take a pre-trained BERT model released by \citet{devlin-etal-2019-bert} and further train it using different masking functions discussed above. 
Like the original BERT framework, we mask the chosen word 80\% of the time, replace it with a random word 10\% of the time, and leave it unchanged 10\% of the time. The original BERT model uses WordPiece tokenization \cite{wordpiece} and masks only tokens. Later whole word masking was introduced, where all the tokens associated with a word are masked. Since our task demands whole word masking, and for a fair comparison, we compare our results with both variants of random masking (whole word masking (WWM) and token masking (TM)).

\begin{table}[bth]
\centering\small
\begin{tabular}{lll}
\hline
\textbf{Dataset} & \textbf{Train} & \textbf{Test} \\
\hline
Amazon Review & 40,000 & 5,000 \\
Movie Review & 8,534 & 2,128 \\\hline
Gab & 28,776 & 5,000 \\
Reddit & 17,314 & 5,000 \\\hline
GYAFC - Family & 103,934 & 2,664 \\
OC & 8,730 & 1,400 \\\hline
Humour & 100,000 & 10,000 \\\hline
BTC & 5,551 & 4,000 \\
WNUT-17 & 3,394 & 2,296 \\
\hline
\end{tabular}
\caption{Dataset Statistics}\label{stats}
\end{table}

We use a small chunk ($\sim$ 100 Mb) of BookCorpus \cite{bookcorpus} and train the BERT model on it for 20k steps using the above masking strategy. For a fair comparison, we also pre-train another BERT on this corpus using random masking (both token masking and whole word masking). To compare our results with \citet{gu2020train}, we also train BERT on the Amazon Reviews dataset using our selective masking and finetune it on the Movie Reviews dataset (discussed below). We test our approach on a variety of downstream tasks. We finetune for ten epochs and report the best result.

\begin{table*}[bth]
\centering\scriptsize
\begin{tabular}{l||lll||ll||ll||l||ll}
\hline
\textbf{Masking Function} & \multicolumn{3}{c||}{Sentiment Analysis} & \multicolumn{2}{c||}{H.S.D}& \multicolumn{2}{c||}{Formality Analysis}& H.D. & \multicolumn{2}{c}{NER}\\

& \textbf{Amazon} &\textbf{MR} &\textbf{MR+A.} & \textbf{Gab} & \textbf{Reddit} & \textbf{OC} & \textbf{Family} & \textbf{ColBERT} & \textbf{BTC} & \textbf{WNUT-17} \\
\hline

\textbf{Random (TM)}& 59.46 & 85.66 & 87.43 & 92.22 & 88.92 & 85.21 & 87.88 & \textbf{98.68} & 49.99 & 18.80 \\

\textbf{Random (WWM)}& 59.79 & 85.47 & 87.05 & 92.26 & 87.88 & 84.86 & 88.25 & 98.45 & 49.93 & 19.25 \\

\textbf{MF 1}& 60.20 & 85.66 & 87.52 & 92.22 & 88.44 & 85.43 & 87.95 & 98.57 & 48.33 & 20.71 \\

\textbf{MF 2}& \textbf{61.42} & 86.58 & 88.34 & 93.14 & \textbf{89.99} & 86.57 & 89.18 & 98.64 & 50.26 & 21.04 \\

\textbf{MF 3}& 60.53 &\textbf{87.60} & \textbf{89.65} & \textbf{93.20} & 89.34 & \textbf{87.43} & \textbf{89.03} & 98.61 & \textbf{51.78} & \textbf{21.96} \\
\hline
\end{tabular}
\caption{Results on different downstream tasks. We report F-1 score for the NER task. MR+A. means Movie Reviews with Selective Masking on Amazon Reviews data, H.S.D. means Hate Speech Detection, H.D. means Humor Detection, and M.F. means Masking Function.}
\label{table:results}
\end{table*}

For \textbf{sentiment classification} task we use Amazon review dataset \cite{amazon} and movie review dataset (referred as MR) \cite{pang-lee-2005-seeing}. Amazon review is a multiclass classification problem, while the movie review dataset is a binary classification. 
For \textbf{hate speech detection} we use data from Gab and Reddit \cite{hatespeech} in a binary classification setup. For \textbf{formality detection} task we use GYAFC dataset \cite{gyafc} and online communication (OC) data \cite{pavlick-tetreault-2016-empirical}. For \textbf{NER on informal text} we use text from social media as a proxy to informal text. To this end, we use the dataset provided by Broad Twitter Corpus (referred as BTC) \citep{derczynski-etal-2016-broad} and WNUT-17 \citep{derczynski-etal-2017-results}. For \textbf{humor detection} task we use the dataset by \citet{colbert}. To also test whether the gain in results is due to the actual selective masking strategy and not any other reason, we deliberately use a word list whose words mismatch with the type of words present in the dataset for the humor detection task. The words present in the word list were the words commonly used in off-color humor (dark sexist, racial, etc. jokes). In contrast, the fine-tuning dataset consisted of short formal text with no vulgarity. We expect that there will be no significant difference in results using selective masking as compared to random masking.
Table \ref{stats} provide statistics about each of these dataset. Note that we report the combined test and validation dataset as our test dataset.

For the sentiment analysis task, we use the word list provided by \citet{slist}. We had 1,856 words in each class. For hate speech detection, we used the list of hate speech words from here\footnote{\url{https://www.cs.cmu.edu/~biglou/resources/bad-words.txt}}. For non-hate words, we used POS-tagging to get positive adjectives and adverbs. We also collect neutral words from Wikipedia. There were 426 words in each class. For formality detection, we used the list provided by Julian Brooke along with words collected from the internet to get a final list with 620 words in each class. For humor detection, we used the word list provided by \citet{hlist} to get 400 words in each class. For NER, we used the same list as Formality detection.

\section{Results and Discussion}
Tables \ref{table:results} show the results for different downstream tasks. We see that tailoring the BERT model using selective masking methodology helps and it outperforms BERT's random masking for 4 out of 5 tasks. As expected, there were no gains in results for Humor Detection because we did not use a suitable word list. Thus, this shows that if there is a mismatch in the words found in the word list and the downstream task, our strategy will provide no gains. Due to architectural limitations, we could not train BERT from scratch using the proposed methodology or try \citet{gu2020train} like three-stage training. However, with just 20k steps, we are able to achieve improvements over random masking, showing the strength of the approach.

We notice that no single masking function performs the best for all the tasks. Masking Function 2 and 3 outperform Random Masking almost all the time. Masking Function 3 achieves the best results 70\% of the time, suggesting the superiority of the exponential functions in capturing the relationship between masking probability and score. For no task, Masking Function 1 performed the best. This confirms the hypothesis of \citet{inbook} and \citet{pavlick-tetreault-2016-empirical}. They argued that sentiments or formality does not always depend on extreme words, and we show it experimentally. To this end, our approach of unsupervised training using word lists and embeddings helps in capturing the domain-specific and task-specific patterns. Which masking function will perform the best depends upon the choice of the word list and downstream task.

Comparing our results with \citet{gu2020train}, they report a gain of 2.14\% (compared to the BERT model) on the Movie Reviews dataset using selective masking and Amazon reviews dataset (in-domain data). We have achieved a gain of  1.94\% (compared to the BERT model) in the accuracy using selective masking alone. When we use in-domain data too, we achieve a gain of  3.99\% compared to the BERT model. Thus our strategy is effective. \citet{gu2020train} cannot work for non-classification tasks. We showed that our approach works for NER as well and thus is more generalizable.



\section{Conclusion}
We presented a framework for tailoring the BERT model for downstream tasks using selective masking. Unlike the previous approach, we use easily accessible word lists instead of the specific in-domain data. Our approach is also more generalizable as it works for non-classification tasks as well. Experimental results on various downstream tasks show that our approach of training BERT using selective masking helps in capturing the domain-specific and task-specific patterns. But the method described in this paper cannot be generalized to all Natural Language Understanding (NLU) tasks. It would be an interesting problem to find a way to incorporate selective masking for other NLU tasks such as generic question answering or translation where there are no clear boundaries for categorizing/scoring the relevance of the words for the task at hand.

\bibliography{anthology,custom}
\bibliographystyle{acl_natbib}

\end{document}